\documentclass[10pt,twocolumn,letterpaper]{article}

\usepackage{cvpr}              

\usepackage{graphicx}
\usepackage{amsmath}
\usepackage{amssymb}
\usepackage{booktabs}
\usepackage{times}
\usepackage{helvet}
\usepackage{courier}
\usepackage{graphicx} 
\usepackage{array}
\usepackage{amsmath}
\usepackage{graphicx}
\usepackage{multirow} 
\usepackage{float}
\usepackage{booktabs}
\usepackage{longtable}
\usepackage{booktabs}
\usepackage{multirow}
\usepackage{tabularx}
\usepackage{array}
\usepackage{caption}
\usepackage{color}
\usepackage{pifont}
\usepackage{colortbl}
\usepackage{graphicx} 
\usepackage{caption} 
\usepackage{soul}
\usepackage{algorithm}
\usepackage{algorithmic}
\usepackage[dvipsnames]{xcolor}
\usepackage{makecell}
\usepackage{adjustbox}
\usepackage[pagebackref,breaklinks,colorlinks]{hyperref}

\usepackage[capitalize]{cleveref}
\crefname{section}{Sec.}{Secs.}
\Crefname{section}{Section}{Sections}
\Crefname{table}{Table}{Tables}
\crefname{table}{Tab.}{Tabs.}

\def\cvprPaperID{*****} 
\def\confName{CVPR}
\def\confYear{2022}

\begin{document}

\title{GoT-CQA: Graph-of-Thought Guided Compositional Reasoning for \\ Chart Question Answering}

\author{Lingling Zhang\textsuperscript{1\#} \and Muye Huang\textsuperscript{1\#} \and Qianying Wang\textsuperscript{2}\thanks{Corresponding author. \# These authors contributed to the work equally.} \and Yaxian Wang\textsuperscript{1} \and Wenjun Wu\textsuperscript{1} \and Jun Liu\textsuperscript{1}\\
Xi’an Jiaotong University\textsuperscript{1} \quad
Lenovo Research\textsuperscript{2}\\
{\tt\small \{huangmuye, wyx1566, nickjun98\}@stu.xjtu.edu.cn} \quad {\tt\small wangqya@Lenovo.com}
\\
{\tt\small \{zhanglling, liukeen\}@xjtu.edu.cn}
}

\maketitle

\begin{abstract}
   Chart Question Answering (CQA) aims at answering questions based on the visual chart content, which
plays an important role in chart summarization, business data analysis, and data report generation.
CQA is a challenging multi-modal task because of the strong context dependence and complex reasoning requirement. 
The former refers to answering this question strictly based on the analysis of the visual content or internal data of the given chart, while the latter emphasizes the various logical and numerical reasoning involved in answer prediction process.
In this paper, we pay more attention on the complex reasoning in CQA task, and propose a novel Graph-of-Thought (GoT) guided compositional reasoning model called GoT-CQA to overcome this problem.
At first, we transform the chart-oriented question into a directed acyclic GoT composed of multiple operator nodes, including localization, numerical and logical operator.
It reflects the human brain's solution process to this question intuitively.
After that, we design an efficient auto-compositional reasoning framework guided by the GoT, to excute the multi-step reasoning operations in various types of questions. 
Comprehensive experiments on ChartQA and PlotQA-D datasets show
that GoT-CQA achieves outstanding performance, especially in complex human-written and reasoning questions, comparing with
the latest popular baselines.
\end{abstract}

\section{Introduction}
\label{Introduction}

Charts and figures are an effective visual expression about data statistics, the trends, and outliers, which exist widely in academic articles, technical reports, and various websites.
Chart question answering (CQA) refers to correctly answer the given question based on the visual chart content.
It plays an important role for a range of downstream tasks, such as chart summarization, business data analysis, and data report generation.
With the development of multi-modal understanding and reasoning techniques, CQA task has received growing attentions in recent years \cite{kafle2020answering,methani2020plotqa,masry2021integrating}.
\begin{figure}[t]
	\centering
	\includegraphics[width=3.2 in]{ 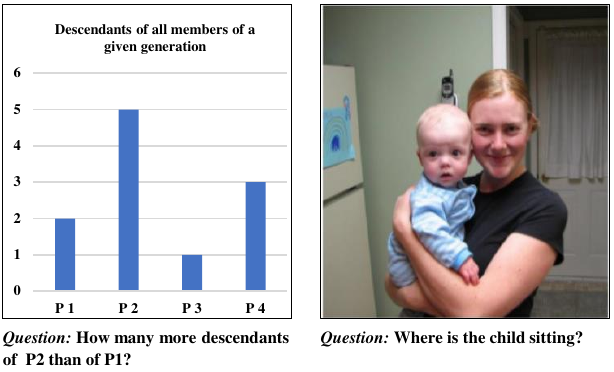}
	\caption{Examples of CQA and VQA task.
		\label{fig:vqa-cqa example}}
\end{figure}

As everyone knows, the hot research visual question answering (VQA) \cite{li2023weakly,wang2023disavr} is closely related to CQA task, where the former is the question answering (QA) task on natural images and the latter focues on the specific charts. 
Although the performance on VQA task has made break-through progress, CQA is still challenging because of the strong context dependence and the complex reasoning requirement.
On one hand, strong context dependence means answering a question strictly rely on the visual and underlying data information in the given chart, rather than the common sense or priors.
Natural images directly reflect the phenomena in real world, thus the common sense or priors mined from previous QA pairs can help better answer the current question.
As shown in right part of Fig. \ref{fig:vqa-cqa example}, 
the answer \ul{``\emph{arms}''} is a very likely candidate for question \ul{``\emph{Where is the child sitting?}''}, even if the model does not understand the image context.
However, for the question \ul{``\emph{How many more descendants of P2 than of P1?}''}, the locations and values of \emph{P1} and \emph{P2} must be extracted from given chart, even if the labels \emph{P1} and \emph{P2} have appeared in other charts.
In this case, the fine-grained parser of charts, including coordinates, legends, point locations, bar heights and other information, is a necessary preliminary work for CQA. 
And current researches \cite{rane2021chartreader,kato2022parsing} generally combine multiple technologies, such as object detection, OCR, word embedding, and expert rules, to achieve this goal.

On the other hand, complex reasoning that involves several logical and arithmetic operations after the detailed chart parser is required for the answer prediction in CQA. 
Most previous researches \cite{methani2020plotqa,kafle2020answering} regard the CQA as a simple classification task, where the output answers are limited in YES/NO or a fixed vocabulary contained all textual elements in the given chart.
However, the general setting, \emph{e.g.} answers are not appearing in the chart called out of vocabulary (OOV), is more common in realistic application.
Thus, many CQA models such as ChartT5 \cite{2023enhanced}, ChartReader \cite{cheng2023chartreader}, and Matcha \cite{matcha}, are designed based on the large pre-trained vision-language models for the general setting, which improve the answering performance effectively but are still limited on the complex reasoning questions and have great limitation on model interpretability.
In this paper, we focus more on better performing complex reasoning in CQA task, and summarize the following two key issues in this reasoning process.

\begin{figure*}[t]
	\centering
	\includegraphics[width=6.4 in]{ 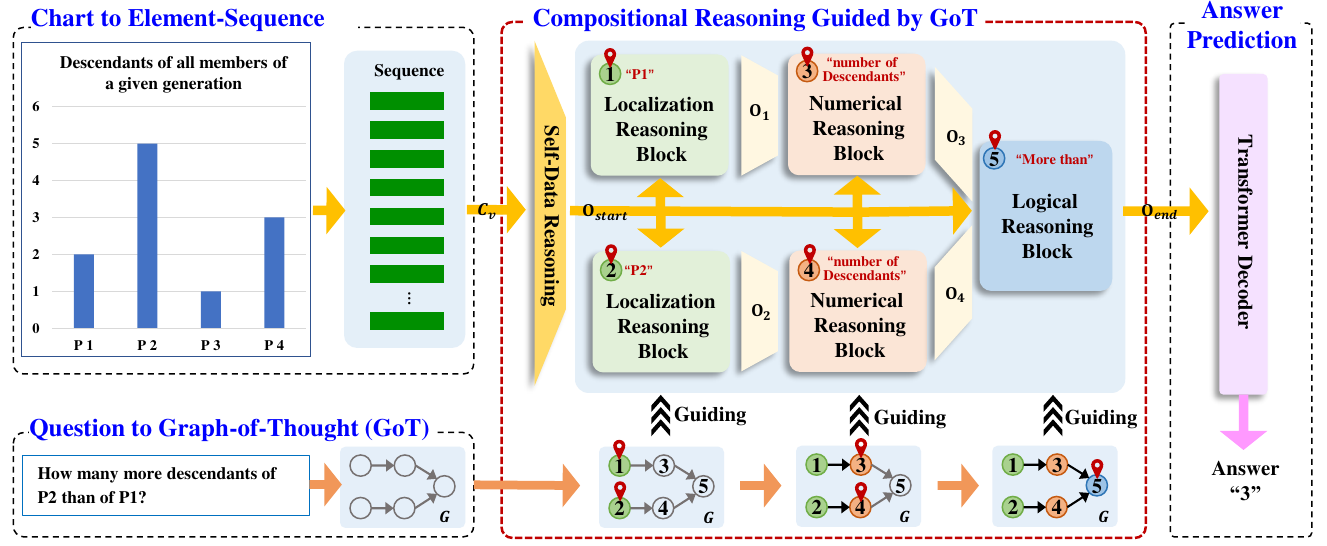}
	\caption{The overview of the proposed GoT-CQA framework. 
 \label{fig:frame}}
\end{figure*}

\textbf{Firstly, how to implement the multi-step numerical and logical reasoning?}
In left part of Fig. \ref{fig:vqa-cqa example},
for question \ul{``\emph{How many more descendants of P2 than of P1?}''}, its reasoning procedure involves finding the bars corresponds to \emph{P1} and \emph{P2} in chart, estimating the descendant number of them, and finally executing the difference operation.
For the more challenging question \ul{``\emph{Are the number of descendants of P2 more than the average descendants?}''}, in addition to identifying the number of bars, the model needs to evaluate the corresponding value of each bar, and perform averaging and comparison operations.
Overall speaking, it is necessary in CQA model to perform multi-step operations such as finding, estimating, comparison, summation, and subtraction.

\textbf{Secondly, how to establish a unified reasoning framework for various types of questions?}
Taking the large-scale dataset PlotQA-D \cite{methani2020plotqa} as an example, the questions are constructed with 74 templates, and are categorized into three groups: structural understanding, data retrieval, or reasoning.
Apparently, different types of questions involve different inference operations, for instance the inference of question \ul{``\emph{How many descendants does P2 have?}''}
is much simpler than
\ul{``\emph{Are the number of descendants of P2 more than the average descendants?}''}.
However, designing a reasoning model for each type of question is not only time-consuming and expensive, but also difficult to generalize to new question types. 
Thus it is essential to design a unified and flexible reasoning framework for various questions.

To address the above issues, we propose a novel \textbf{G}raph-\textbf{o}f-\textbf{T}hought (GoT) guided compositional reasoning framework for the challenging CQA, which is called GoT-CQA model.
GoT-CQA includes three modules: chart \& question parsing, compositional reasoning, and answering module.
Chart \& question module applies the pre-trained large-scale models to generate an feature sequence for given chart and a GoT corresponds to the question, where the directed acyclic GoT reflects the types and orders of the reasoning operations need to be performed.
Compositional reasoning module achieves complex reasoning over the chart data-flow under the guidance of GoT.
This module designs the self-data reasoning block and localization, numerical and logical reasoning operation blocks, and the reasoning framework is automatically formed by an orderly combination of these blocks according to the question's GoT. 
The answering module is a general transformer decoder to jointly generate the YES/NO, in-vocabulary, or OOV answers.
Our main contributions can be summarized into four folds:
\begin{itemize}
	\item We reveal the strong context dependence and complex reasoning requirement for the challenging CQA task. And we propose the novel model GoT-CQA as the first attempt to solve this complex reasoning problem in CQA.
	\item We transform the question about chart into an interesting directed acyclic GoT to guide the answer reasoning.
	GoT splits the complex reasoning process into several ordered localization, numerical, or logical operations. 
	\item We present an efficient automatic compositional reasoning pattern that generalizes to various types of chart-oriented questions.
	And it enhances the reasoning interpretability to a certain extent.
	\item We conduct extensive experiments on dataset ChartQA and PlotQA-D to verify the superiority of GoT-CQA. The results
	show that GoT-CQA achieves a good improvement, especially in complex reasoning questions.
\end{itemize}

\section{Methodology}
\label{Methodology}
Fig. \ref{fig:frame} illustrates the overall framework of GoT-CQA, and it includes three  modules: 1) Chart \& question parsing module (Left Part) to extract the visual feature sequence from chart and generate a GoT corresponds to the question;
2) Compositional reasoning module (Center Part) to perform the complex localization, numerical, and logical reasoning over chart's data flow guided by GoT;
3) Answering module (Right Part) to generate the target answer based on reasoning result.
Details of these modules are introduced as follows.

\subsection{Chart \& Question Parsing Module}

\noindent\textbf{Chart to Sequence-Feature.} 
Following to the work \cite{unichart}, we employ the pretrained OCR-free architecture Donut \cite{kim2022ocr} as the chart encoder.
Donut is originally designed for document image (\emph{e.g.}  receipts) understanding, and the work \cite{unichart} testifies its effectiveness on chart parsing.
For any chart $C$, Donut generates an embedding sequence $\textbf{C}_v$, following the order from the
top-left corner to the bottom-right corner of this chart.

\begin{figure}[t]
	\centering
	\includegraphics[width=3.4 in]{ 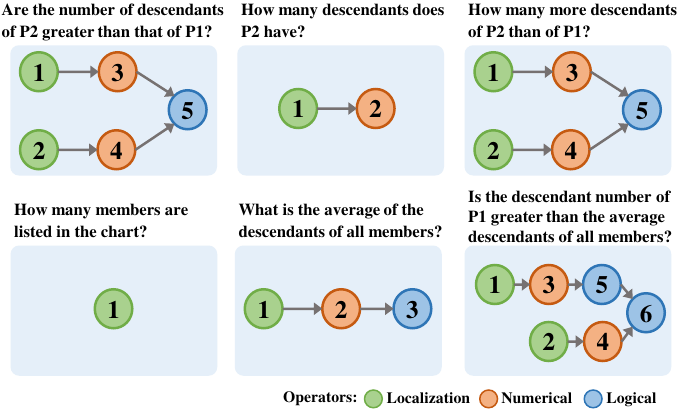}
	\caption{Some GoT examples of questions.
		\label{fig:GoT_example}}	
\end{figure}

\noindent\textbf{Question to Graph-of-Thought.}
For any question $Q$, we find its logic structure is relatively clear and easy to dig out.
To achieve this goal, we first define the following three types of unit operators: \textcircled{1} \textbf{Localization operator (Loc)} means  query the position information for the given input;
\textcircled{2} \textbf{Numerical operator (Num)} means retrieve or simply reasoning the value for the target variable;
And \textcircled{3} \textbf{Logical operator (Log)} means execute some logical inference on multiple values, such as comparison, summation, and maximum.
Apparently, any question can be decomposed into the ordered operations of the above three unit operators.
Namely question $Q$ can be represented as a directed acyclic graph $\mathcal{G}$ composed of multiple operator nodes, which is called Graph-of-Thought (GoT).
This procedure can be formalized as:
\begin{align}
\mathcal{G} & = \{ \mathcal{O}, \mathcal{E}\}, \nonumber\\ 
\mathcal{O} &= \{o_1, o_2, \cdots, o_n\}, o_i = (\widetilde{o}_i, \text{type}(o_i)),  \nonumber\\ 
\mathcal{E} & = \{e_{ij}=(o_i,o_j)\}\subset \mathcal{O}\times \mathcal{O},
\end{align}
where $\mathcal{O}$ is the operator node set, and $n$ records the total number of operators for question $Q$.
For each node $o_i$, it is denoted to a tuple  consisting of the operation content $\widetilde{o}_i$ and  $\text{type}(o_i)\in\{\text{Loc}, \text{Num}, \text{Log}\}$. 
$\mathcal{E}$ is the edge set, and the edge $e_{ij}$ exists when there is a chain of thought from node $o_i$ to $o_j$. 
After analyzing some popular CQA datasets, we found that GoTs can be extracted by pre-defined template rules or the prompted large-scale language models such as GPT.
Fig. \ref{fig:GoT_example} shows several examples of question GoTs.
For question \ul{``\emph{How many more descendants of P2 than of P1?}''}, its GoT includes two localization nodes, two numerical nodes, one logical nodes, and four edges between them. 
The edge $(o_1,o_3)$ indicates that the value of \emph{P1} can only be measured after finding the location of \emph{P1} in chart.
Similarly, edges $(o_3,o_5)$ and $(o_4,o_5)$ indicate that the values of \emph{P1}  and \emph{P2}  need to be obtained before the subtraction operation.  

\subsection{Compositional Reasoning Module}
\begin{figure}[t]
	\centering
	\includegraphics[width=3.2 in]{ 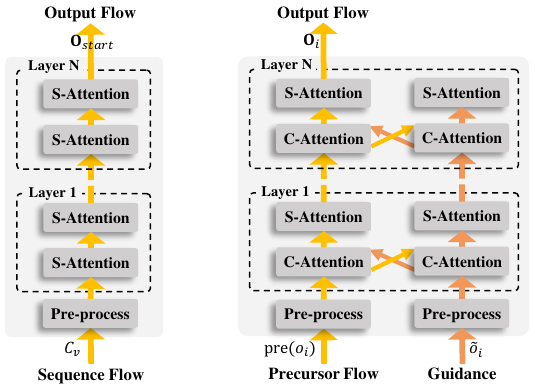}
	\caption{Architecture for four types of reasoning blocks. Left: self-data reasoning; Right: Loc/Num/Log reasoning.
		\label{fig:block}}	
\end{figure}

Given chart data-flow $\textbf{C}_v$, this module accomplishs auto-compositional reasoning under the guidance of GoT $\mathcal{G}$. 
In Fig. \ref{fig:block}, we design four blocks for self-data reasoning and three types of operation (\emph{i.e.} $\{\text{Loc}, \text{Num}, \text{Log}\}$) reasoning.

\noindent\textbf{Self-Data Reasoning.} This block takes visual sequence $\textbf{C}_v$ as input, and performs self-reasoning on the chart to mine the meaning of elements and the relationship between them.
As shown in left part of Fig. \ref{fig:block}, the pre-process layer is a fully connected layer over $\textbf{C}_v$. 
Each reasoning layer is designed two encoders with self-attentions, where each encoder computes the query, key, and value, followed by feed-forward, skip connection and normalization. 
After N-layer, the output of this block is denoted as the feature matrix $\mathbf{O}_{start}$, which is applyed to the following compositional reasoning.

\noindent\textbf{Loc/Num/Log Reasoning.} 
In right part of Fig. \ref{fig:block}, the architectures of localization, numerical, and logical reasoning blocks are similar.
Note that each operator node $o_i$ in GoT corresponds to a reasoning block.
And its input consists of two parts:
the current guidance information $\widetilde{o}_i$, and the chart data flows  from $o_i$'s precursor nodes $\text{pre}(o_i)$.
This procedure is formalized to Eq. (\ref{block_fun}) with the output data flow matrix $\mathbf{O}_i$,
and the node set $\text{pre}(o_i)$ is arrived at by
Eq. (\ref{pre}). 
\begin{align}
	\label{block_fun}
	\mathbf{O}_i = 
	\text{Block}_{\text{type}({o_i})}\left(\text{pre}(o_i),\widetilde{o}_i\right), \ \ \ \ \ \ \ \ \ \ \ \ \ \ \ \ \ \ \ \ \ \\
	\label{pre}
	\text{pre}(o_i) = 
	\begin{cases}
	\{o_{start}, o_k;e_{ki}\in \mathcal{E}\},\ &\text{indegree}(o_i)\neq0\\
	\{o_{start}\}, &\text{otherwise}
	\end{cases}
\end{align}
where the function $\text{indegree}(\cdot)$ computes the in-degree value of the input operator node.
$o_{start}$ is a virtual node outside the GoT, and it stores the original chart flow data matrix $\mathbf{O}_{start}$ that obtained by self-data reasoning block.
Especially, if the in-degree of node $o_i$ is zero, it is the starting node of compositional reasoning with initial data flow $\mathbf{O}_{start}$ as input.
There is also a pre-process layer in Loc/Num/Log reasoning block.
On the one hand, this layer fuses all the precursor data flows $\{\mathbf{O}_k;o_k\in  \text{pre}(o_i)\}$ into a new feature matrix. 
On the other hand, it uses the pre-trained language model (\emph{e.g.} Bert) to extract  the semantic feature of textual guidance information $\widetilde{o}_i$.
For each reasoning layer, there are two types of encoders: self-attention and cross-attention.
The architecture of the former is the same as in the self-data reasoning block.
The latter is designed to interact the information between the guidance and the chart data flow, where the transformer's keys and values are mutually exchanged resulting
a cross-attention.
After N-layer, the output data flow $\mathbf{O}_i$ is the input of the reasoning blocks of $o_i$'s successor nodes.

\noindent\textbf{Compositional Reasoning Guided by GoT.}
For any question $Q$, 
its reasoning network is formed by an orderly combination of localization, numerical, logical blocks under the guidance of GoT $\mathcal{G}$.
This is a novel auto-compositional reasoning process, which means that the architecture of reasoning network are varies for different questions.
After the construction, all reasoning blocks are executed in multiple steps.
For each step, these blocks whose related precursor nodes 
have been inferred in the current state, are activated and executed.
Taking the question \ul{``\emph{How many more descendants of P2 than of P1?}''} as an example, we first execute two localization blocks to find  the location of \emph{P1} and \emph{P2}, then execute numerical blocks for estimating their values, and finally execute logical block for subtraction operation.
The compositional reasoning process of the above CQA example is shown in the center of Fig. \ref{fig:frame}.
Specially, the data flow output by the node whose out-degree is zero in GoT $\mathcal{G}$, is the result of the reasoning module.
It provides the evidence for answering module in the next section.
We found that there is generally only one such terminal node in the GoT, and  the its output is denoted $\mathbf{O}_{end}$ for convenience.

\subsection{Answering Module}
For question $Q$, its answer may be YES/NO, a textual element in chart, or the result of some numerical operation.
Following the latest CQA models \cite{matcha,unichart},
We  design a simple  transformer decoder that takes data flow $\mathbf{O}_{end}$ as input and generates the final answer for all types of questions.
The traditional cross-entropy loss is applied to train this model.

\section{Experiment}
\label{Experiment}

\subsection{Experimental setting}
\label{Experimental setting}
\textbf{Datasets.} 
(1) \textbf{ChartQA} \cite{masry2022chartqa} is constructed with real-world charts and human-authored question-answer pairs.
It covers 9.6K human-written questions focusing on logical and visual reasoning, and 23.1K
questions generated from human-written chart summaries. 
(2) \textbf{PlotQA-D} \cite{methani2020plotqa} is another large-scale dataset that includes two benchmarks called PlotQA-D1 and PlotQA-D2.
Different from ChartQA, the charts in PlotQA-D are generated with a programming tool and the questions are created with 74 pre-defined templates that can be grouped into structural understanding, data retrieval, and reasoning.

\textbf{Competitors.}
Nine latest works on CQA task, including PlotQA-M \cite{methani2020plotqa}, PReFIL \cite{kafle2020answering}, CRCT \cite{levy2022classification}, VisionTapas \cite{masry2022chartqa}, ChartT5 \cite{2023enhanced}, Pix2Struct \cite{lee2023pix2struct}, 
ChartReader \cite{cheng2023chartreader},
MatCha \cite{matcha},
and UniChart \cite{unichart}, are applied to compare with our model.
Note that we directly run their publicly available codes or use the published evaluation results on ChartQA and PlotQA-D.
\begin{table}[t]
	\centering
	\center
	\renewcommand{\arraystretch}{1.0}
	\tabcolsep=1 pt
	\begin{tabular}{cccc}
		\toprule
		Model
		&Human &Augmented & Overall \\  \midrule 
		VisionTapas (ACL-Findings2022) &29.6 &61.4 &45.5\\
		ChartT5 (ACL-Findings2023) &31.8 &74.4 &53.1\\
		Pix2Struct (ICML2023) &30.5 &81.6 &56.0 \\
		ChartReader (ICCV2023) &-  &-  &52.6 \\
		MatCha (ACL2023) &38.2  &90.2  &64.2 \\
		UniChart (EMNLP2023) &43.9 &87.8 &65.8  \\
		\rowcolor{Gray!35} GoT-CQA (Ours)  &\textbf{47.1}	&\textbf{87.9}	&\textbf{67.5} 	 \\
		\bottomrule      
	\end{tabular}
 \caption{Comparison results on ChartQA dataset. 
 	\label{tab1:result}}
\end{table}
\begin{table*}[t]
	\centering
	\center
	\renewcommand{\arraystretch}{1.0}
	\tabcolsep=11 pt
	\begin{tabular}{ccccccccc}
		\toprule
		\multicolumn{1}{c}{\multirow{2}{*}{Model}} & \multicolumn{4}{c}{PlotQA-D1}                                   & \multicolumn{4}{c}{PlotQA-D2} 
		\\ \cmidrule(r){2-5} \cmidrule(r){6-9} 
		& S     & D     & R     & Overall & S     & D     & R   & Overall \\  \midrule 
		PlotQA-M (WACV2020)	 & 86.3 & 45.7 & 31.2 & 54.0   & 76.0 & 58.9 & 15.8  & 22.5\\
		PReFIL (WACV2020) & 96.7 & 58.7 & 31.7 & 57.9   & 96.7 & 21.9  & 3.9    & 10.4 \\
		CRCT (ECCV2022) & 96.1 & 94.5 & 54.9 & 76.9   & 96.2 & 66.7 & 25.8  & 34.4  \\
		VisionTapas (ACL-Findings2022) &-  &-  &- &65.3 &-  &-  &-   &42.5\\
		Pix2Struct (ICML2023) &-  &-  &- &73.2 &-  &-  &-   &71.9\\
		ChartReader (ICCV2023) &-  &-  &- &78.1 &-  &-  &-   &59.3\\
		MatCha (ACL2023) &-  &-  &- &92.3 &-  &-  &-   &\textbf{90.7}\\
	    \rowcolor{Gray!35} GoT-CQA (Ours)  &\textbf{98.4}	&\textbf{97.8}	&\textbf{86.5} 	&\textbf{92.8} 	&\textbf{98.2}	&\textbf{88.8}	&\textbf{72.6}	
		&{78.3} \\
		\bottomrule      
	\end{tabular}
	\caption{Comparison results on PlotQA-D dataset. The ``S'', ``D'', and ``R'' rows record the performance on structural, data retrieval, and reasoning type questions, respectively.
		\label{tab2:result}}
	
\end{table*}

\textbf{Settings.}
In question parsing module, GoTs of questions in ChartQA are generated by large-language model Qwen2-7B\footnote{https://https//qwenlm.github.io/blog/qwen2/wen2/}, while the GoTs in PlotQA are obtained with a template expert library.
In compositional reasoning module, the layer of self-data reasoning and the Loc/Num/Log reasoning blocks are set to 4 and 1, respectively.
And in answering module, the 12-layer transformer decoder framework from translation model mBART \cite{2020Multilingual} is applied.
During training, the pre-trained parameters from Donut encoder are applied to initialize our GoT-CQA.
For ChartQA dataset, similar to \cite{unichart}, we conduct a first-stage pre-training on a extra chart dataset for 200k steps, followed by a second-stage fine-tuning on ChartQA dataset for 5 epochs. For PlotQA dataset, we perform a single-stage training on the PlotQA-D1 and PlotQA-D2 datasets for 200k steps. 
All experiments are completed on two A100 GPUs.

\begin{table*}[t]
	\centering
	\center
	\renewcommand{\arraystretch}{1.0}
	\tabcolsep=4.0 pt
	\begin{tabular}{clcccccccc}
		\toprule
		\multicolumn{2}{c}{\multirow{2}{*}{Model}} & \multicolumn{3}{c}{ChartQA}                                   & \multicolumn{4}{c}{PlotQA-D1} 
		&\multicolumn{1}{c}{\multirow{2}{*}{Time cost (Min)}}  
		\\
		\cmidrule(r){3-5} \cmidrule(r){6-9} 
		&  &Human &Augmented  & Overall   &S &D &R & Overall   \\  \midrule 
		\multirow{3}{*}{w/o GoT} &One operator &44.7 &87.8 &66.2 &96.9 &96.3 &84.3 &90.9 &41.8
		\\
		&Two operators (Find+Log) &45.0 &87.0 &66.0 &97.9 &96.0 &84.1 &91.1 &66.3
\\
		&Three operators (Loc+Num+Log) &45.2 &87.3 &66.3 &98.0 &97.4 &84.2 &91.7 &79.2\\
		\hline
		\multirow{2}{*}{w/ GoT}
		&Two operators (Find+Log) &45.8 &87.0 &66.4 &98.1 &97.9 &84.4 &91.8 &59.4\\
		 &\cellcolor{Gray!35}Three operators (Loc+Num+Log) &\cellcolor{Gray!35}\textbf{47.1} & \cellcolor{Gray!35}\textbf{87.9} & \cellcolor{Gray!35}\textbf{67.5} & \cellcolor{Gray!35}\textbf{98.4} & \cellcolor{Gray!35}\textbf{97.8} & \cellcolor{Gray!35}\textbf{86.5} &\cellcolor{Gray!35}\textbf{92.8}
		 &\cellcolor{Gray!35}60.9
		\\
		\bottomrule      
	\end{tabular}
	\caption{Ablation study on GoT guided compositional reasoning. Values achieved by GoT-CQA are highlighted in gray. \label{tab:GoT_ablation}}
\end{table*}

\subsection{Performance Comparison}
\label{Performance Comparison}
 \textbf{Results on ChartQA.} Table \ref{tab1:result} reports the answer scores over human and augmented questions from ChartQA.
From these results, we make the following two observations.
(1) our GoT-CQA consistently perform better than other competitors, in addition to the ``Augmented'' score obtained by  MatCha.
And GoT-CQA's advantage on ``Human'' questions is more prominent, which is nearly 3.2\% higher than the best comparison result.
This indicates that the proposed GoT guided compositional reasoning is beneficial to CQA task, especially for the challenging human-written questions.
(2) For all methods, the performance scores on ``Human'' questions are significantly worse than those on ``Augmented'' questions.
This is reasonable because the human annotated questions are more complex and require the more challenging reasoning compared to these augmented questions.

\textbf{Results on PlotQA-D.} 
Table \ref{tab2:result} reports the scores over structural, data retrieval and reasoning questions, and the overall questions from PlotQA-D1 and PlotQA-D2 test set.
The results are very consistent with those from  ChartQA. 
(1) GoT-CQA still achieves the performance improvement on all test data,  in addition to the overall performance on PlotQA-D2 achieved by MatCha.
This phenomenon further demonstrates the outstanding ability of GoT-CQA on solving the more challenging questions.
Note that the competitor MatCha achieves the outstanding performance on PlotQA-D2. 
This is because MatCha is pre-trained over the large scale MATH  and DROP datasets, which significantly enhances its mathematical calculation abilities \cite{matcha}.
(2) With all methods, the performance scores worsen when the question type changes from structural to data retrieval, then to reasoning. And the answer performance of current models on the reasoning questions is less than 87\%.
This is because the structural questions just focus on chart’s structure; data retrieval questions require to search for relevant information from the chart; and reasoning questions involve deep reasoning based on chart data understanding. 

\begin{table*}[t]
	\centering
	\center
	\renewcommand{\arraystretch}{1.0}
	\tabcolsep=14 pt
	\begin{tabular}{clcccccc}
		\toprule
		\multicolumn{1}{c}{\multirow{2}{*}{Settings}} & \multicolumn{3}{c}{ChartQA}                                   & \multicolumn{4}{c}{PlotQA-D1}  
		\\
		\cmidrule(r){2-4} \cmidrule(r){5-8} 
		&Human &Augmented  & Overall   &S &D &R & Overall   \\  \midrule 
		\multicolumn{7}{l}{\emph{Layer number of Self-data Reasoning}}   \\
		2   &44.3 &86.8 &65.6 &98.5 &97.6 &85.0 &92.1
 		\\ 
		3   &46.3 &87.0 &66.7 &98.4 &97.4 &85.8 &92.4
\\
		\rowcolor{Gray!35} 4   &\textbf{47.1} &\textbf{87.9} &\textbf{67.5} &98.4 &\textbf{97.8} &\textbf{86.5} &\textbf{92.8}\\
		5   &45.6 &87.0 &66.3 &\textbf{99.0} &97.4 &85.6 &92.5
\\
		
		\midrule
		\multicolumn{7}{l}{\emph{Layer number of Loc/Num/Log Reasoning}}   \\
		\rowcolor{Gray!35} 1  &\textbf{47.1} &\textbf{87.9} &\textbf{67.5} &98.4 &\textbf{97.8} &\textbf{86.5} &\textbf{92.8}
 \\
		2 &45.3 &87.3 &66.3 &98.4 &97.5 &85.4 &92.2 \\
		3  &44.2 &87.2 &65.7 &\textbf{99.0} &97.5 &84.6 &92.0 \\
		\midrule
		\multicolumn{7}{l}{\emph{Architecture of Loc/Num/Log Reasoning}}  \\
		\rowcolor{Gray!35} SELF-CROSS  &\textbf{47.1} &\textbf{87.9} &\textbf{67.5} &\textbf{98.4} &\textbf{97.8} &\textbf{86.5} &\textbf{92.8}\\
		CROSS-CROSS  &45.4 &87.5 &66.5 &98.3 &98.0 &84.4 &91.8
\\
		SELF-SELF  &45.8 &87.4 &66.6 &97.9 &97.0 &83.6 &91.1

 \\
		\bottomrule      
	\end{tabular}
	\caption{Ablation study on GoT guided compositional reasoning. Values achieved by GoT-CQA are highlighted in gray. \label{tab:block_arch}}
\end{table*}

\begin{table*}[h]
	\centering
	\center
	\renewcommand{\arraystretch}{0.1}
	\tabcolsep=.5 pt
	\begin{tabular}{p{4.8cm}<{\centering}llcccc}
		\toprule
		Charts &Questions & Types & Ground-Truth & MatCha &UniChart & GoT-CQA  \\  \midrule 
		\multirow{5}{*}{
			\begin{minipage}[b]{4.7cm}
				\centering
                    \vspace{-0.6cm}
				\raisebox{-.5\height}{\includegraphics[width=4.0cm, height=4.0cm]{  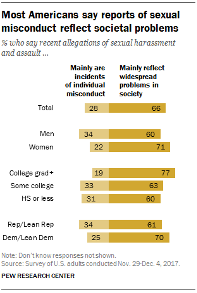}}
			\end{minipage}
		} 
		&\makecell[l]{\emph{\textbf{Q1.1:}} How many values are \\below 30 in Mainly are incidents \\of individual misconduct?} & \makecell[l]{Human}
		&4 &3 &2 &\cellcolor{yellow!20}\textbf{4}  \\ \cmidrule{2-7}
		&\makecell[l]{\emph{\textbf{Q1.2:}} What value is been shown \\twice in the Mainly reflect \\widespread problems in society?}
        &\makecell[l]{Human}	&60 &0.66 &\cellcolor{yellow!20}\textbf{66}  &\cellcolor{yellow!20}\textbf{66} 	
 \\ \cmidrule{2-7}
		& \makecell[l]{\ \ }\\ 
        & \makecell[l]{\ \ }\\ 
        & \makecell[l]{\ \ }\\ 
		&
		\\
		\midrule 
		\multirow{4}{*}{
			\begin{minipage}[b]{4.7cm}
				\centering
                    \vspace{-0.4cm}
				\raisebox{-.5\height}{\includegraphics[width=4.0cm, height=3.1cm]{  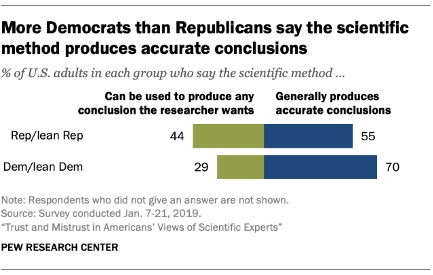}}
			\end{minipage}
		} 
		&\makecell[l]{\emph{\textbf{Q2.1:}} What is the ratio between \\the last bar (dem/lean dem)?} &\makecell[l]{Human} &0.414 &0.214 & 0.168 &\cellcolor{yellow!20}\textbf{0.418}\\ \cmidrule{2-7} 
		&\makecell[l]{\emph{\textbf{Q2.2:}} The blue bar represents \\what (value 55,70)?}
		&\makecell[l]{Human} &\makecell[c]{Generally \\produces \\accurate \\conclusions}
        &1.6 &YES & No \\ \cmidrule{2-7} 
		& \makecell[l]{\ \ }\\ 
        &
		\\
		\midrule 
		\multirow{3}{*}{
			\begin{minipage}[b]{4.7cm}
				\centering
                    \vspace{-0.4cm}
				\raisebox{-.5\height}{\includegraphics[width=4.5cm, height=3.4cm]{  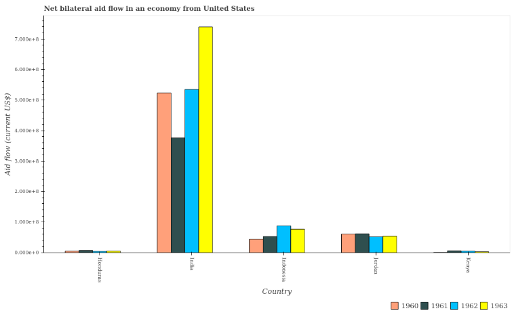}}
			\end{minipage}
		} 
		&\makecell[l]{\emph{\textbf{Q3.1:}} Where does the legend \\appear in the graph?} & \makecell[l]{Structural} &\makecell[c]{Bottom \\Right} &\cellcolor{yellow!20}\textbf{\makecell[c]{Bottom \\Right}} &--- &\cellcolor{yellow!20}\textbf{\makecell[c]{Bottom \\Right}} \\ \cmidrule{2-7}
		&\makecell[l]{\emph{\textbf{Q3.2:}} In how many countries, is \\net bilateral aid flow in 1960 \\greater than the average ...?}
		&\makecell[l]{Reasoning} & 1 &\cellcolor{yellow!20}\textbf{1} &--- &\cellcolor{yellow!20}\textbf{1}   \\  \cmidrule{2-7}
		&\makecell[l]{\emph{\textbf{Q3.3:}} What is the difference \\between the highest and second \\highest net ... in 1961?} &\makecell[l]{Reasoning} &3.23$\times 10^8$ &2000 &--- &\cellcolor{yellow!20}{\textbf{3.14$\times$10$^8$}}
		\\
		\midrule 
		\multirow{3}{*}{
			\begin{minipage}[b]{4.7cm}
				\centering
                    \vspace{-0.4cm}
				\raisebox{-.5\height}{\includegraphics[width=4.5cm, height=3.0cm]{  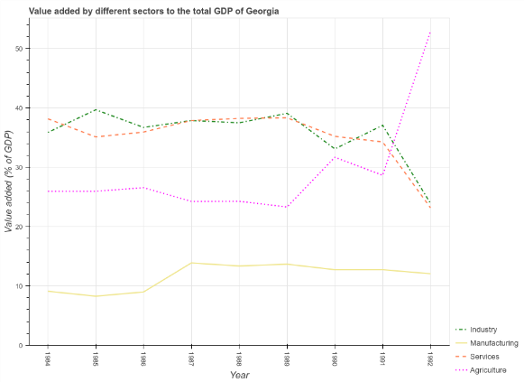}}
			\end{minipage}
		} 
		&\makecell[l]{\emph{\textbf{Q4.1:}} Does the value added by \\... increase over the years?} & \makecell[l]{Data \\
			Retrieval} &NO &\cellcolor{yellow!20}\textbf{NO} &--- &\cellcolor{yellow!20}\textbf{NO}\\ \cmidrule{2-7}
		&\makecell[l]{\emph{\textbf{Q4.2:}} In the year 1989, what is \\the difference between the value \\added by manufacturing ...?}
		&\makecell[l]{
			Reasoning} & $-$9.64 &\cellcolor{yellow!20}\textbf{$-$9.89} &--- &\cellcolor{yellow!20}\textbf{$-$9.66}   \\  \cmidrule{2-7}
		&\makecell[l]{\emph{\textbf{Q4.3:}} What is the average value \\added by industrial sector per year?} &\makecell[l]{
			Reasoning} &35.65 &21.0 &---&\cellcolor{yellow!20}\textbf{35.74}
		\\ 
		\bottomrule      
	\end{tabular}
	\caption{Case study on ChartQA and PlotQA-D datasets. The correct or optimal results are highlighted in yellow. \label{tab:case}}
\end{table*}
\subsection{Analysis on GoT Guided Compositional Reasoning}

\textbf{GoT Ablation Studies.}
As shown in Table \ref{tab:GoT_ablation}, we conduct a thorough ablation
study on the GoT guided compositional reasoning, to analyze the contribution of key strategies in the proposed
GoT-CQA. 
Here, ``w/'' and ``w/o'' GoT represent whether to utilize the graph-of-thought to guide the answer reasoning, respectively.
Namely, the variant ``w/o'' GoT regards the sentence embedding of entire question as the guidance information.
In addition, the compositional reasoning module includes three types of reasoning blocks corresponds to localization, numerical, logical operators in GoT.
To verify the effectiveness of the operator setting, we combine them together as a model variant, and also combine the localization and numerical operators into the finding (Find) operator as another model variant.
According to the results in Table \ref{tab:GoT_ablation}, we observe the three points.
(1) Whether setting two or three operators, the overall performance of model variant w/o GoT is lower than model GoT-CQA. 
(2) Compared to one or two operators, the model GoT-CQA with three operators (Loc+Num+Log) could achieve the comparable performance on ``Augmented", ``S", and ``D"  questions, but it is more outstanding on ``Human" questions from ChartQA and ``Reasoning'' questions from PlotQA-D1. 
(3) The model with three operators does not cause too much time overhead.
This benefits from the proposed auto-compositional reasoning pattern.
That is to say, for the complex question, its structure of GoT corresponds to more localization, numerical, and reasoning operators, thus its inference time is longer; for simple question, it still requires fewer blocks and time cost.

\textbf{Reasoning Architecture Studies.}
In this section, we analyze the block architecture in compositional reasoning module from the following three aspects.

\emph{1) Analysis on Self-Data Reasoning Layers.} 
As shown in the upper part of Table  \ref{tab:block_arch}, we vary the layer number of self-data reasoning block,
and record the performance scores on test set of ChartQA and PlotQA-D1.
We observe that the GoT-CQA model achieves satisfactory results with 4-layer self-data reasoning.
Moreover, the shallow layers are insufficient for comprehensive self-data understanding, and the deep layers likely to cause the overfitting problem.

\emph{2) Analysis on Loc/Num/Log Reasoning Layers.} 
As shown in the center part of Table \ref{tab:block_arch}, we 
set the layer number of Loc, Num, and Log reasoning blocks to be the same, and vary it in range $[1,2,3]$ to record the GoT-CQA's question-answering performance on two datasets. 
The experimental results show that a Loc/Num/Log operator does not require a complex multi-layer reasoning block to implement, and one-layer of attention mechanism could obtain good performance on various type of questions.
Moreover, as the number of layers increases, the compositional reasoning module cannot handle the simple classification questions well.

\emph{3) Analysis on Loc/Num/Log Reasoning Architecture.}
As shown in the bottom part of Table \ref{tab:block_arch}, we analyze the effect of different attention mechanisms in reasoning blocks on the CQA performance.
In addition to the cross-stacking mechanism of self- and cross- attentions (\emph{i.e.} SELF-CROSS),
the pure-stacking mechanisms of cross-attention or self-attention (\emph{i.e.} CROSS-CROSS or SELF-SELF) are also implemented in the experiment.
From these results, we observe that the GoT-CQA model with SELF-CROSS strategy is a better choice, because both internal and interactive information mining of question guidance and chart data flow are required in the reasoning procedure.
 
\subsection{Case Study}
\label{Case Study}
To obtain a more insightful understanding of GoT-CQA model, we demonstrate some case studies in Table \ref{tab:case}. 

\textbf{Human Questions from ChartQA: } including cases Q1.1, Q1.2, Q2.1, and Q2.2.
Obviously, GoT-CQA performs better than the popular competitor MatCha and UniChart, over the human annotated questions.
We also notice that the predicted results of all models are wrong on case Q2.2, possibly due to the limited performance of visual feature extraction with pretrained Donut or the self-data reasoning block.

\textbf{Structure or Data Retrieval Questions from PlotQA-D: } including cases Q3.1 and Q4.1.
We conclude that the performance of MatCha and GoT-CQA are both satisfactory on these questions. 
This is because these questions pay more attention on  visual elements or their relations in charts, and they does not involve complex reasoning or calculations.

\textbf{Reasoning Questions from PlotQA-D: } including cases Q3.2, Q3.3, Q4.2, and Q4.3.
Apparently, the predicted results with our GoT-CQA model are closer to the ground-truth compared to MatCha.
This indicates that our compositional reasoning pattern guided by GoT is indeed suitable for parsing and solving the challenging reasoning questions.

\section{Related Work}
\label{Related work}
 \textbf{Chart-to-Table.} It aims to identify the constituent components (\emph{e.g.} bars and legends) in given chart, and extract its underlying data.
Traditional methods \cite{jung2017chartsense,liu2019data} on this task rely on various heuristic rules which do not work well for new chart types.
ChartOCR \cite{luo2021chartocr}  combines the advantages of deep-learning and hand-designed rules to achieve outstanding performance on bar, pie, and line charts.
Recently, ChartReader \cite{cheng2023chartreader}  introduces a transformer-based chart component detection module and an extended pre-trained vision-language model, which achieves SOTA performance on chart-to-table.
Intuitively, it is a reasonable CQA pipeline by combining the chart-to-table and table question answering together. 
However, it not only faces the serious time-consuming annotation issue in chart-to-table stage, but also leads to error accumulation.
Moreover, answering questions does not necessarily require understanding the whole underlying data in chart.

\textbf{Chart-to-Text.}
It aims to generate natural language captions or summaries from the chart image or chart metadata.
Current researches \cite{chen2019figure,qian2021generating,kanthara2022chart} generally design a deep generation model to achieve this model.
For example, Qian \emph{et al.} \cite{qian2021generating} formulated the chart-to-text task as a controlled captioning problem, where the deep model FigJAM is proposed by utilizing metadata
information and a joint static and dynamic dictionary.
Recently, Huang \emph{et al.} \cite{huang2023summaries} pointed that the chart captioning in articles can be solved by summarizing the paragraphs mentioning the chart, and the popular pre-trained language model is applied in the baseline model.
Nonetheless, these methods still could not reflect high-level meanings such as data trends in text generation.

\textbf{Chart Question Answering.}
It aims to answer questions related to charts by mining the visual and textual information.
\cite{kahou2017figureqa} pioneered a synthetic CQA corpus called FigureQA, and 
designed a benchmark baseline with the relation network to output binary (Yes/No) answers.
\cite{kafle2018dvqa}
introduced the synthetic DVQA dataset that contains three forms of CQA: structural, data retrieval, and reasoning.
To avoid biases of synthesized data,
 \cite{chaudhry2020leaf} provided a more complex dataset LEAF-QA in which charts are extracted from varied real-world data sources.
The above researches regard CQA as a simple classification task.
They assume the answer comes from a ﬁxed size vocabulary or is one textual element within the chart.
To the best of our knowledge, PlotQA-D \cite{methani2020plotqa} is the first dataset that includes the more challenging regression scenarios, namely the answers are not presented in chart and need more complex numerical reasoning.
After that, 
\cite{levy2022classification} proposed a joint classification-regression model called CRCT for CQA, which achieved the outstanding results on PlotQA.
Besides, latest works including ChartReader \cite{cheng2023chartreader}, MatCha \cite{matcha}, UniChart \cite{unichart}, apply large pre-trained vision-language models such as BERT \cite{devlin2018bert} and GPT-3 \cite{brown2020language} for CQA task, and they collect massive chart data to train the model parameters.
Nonetheless, the current performance of CQA is far from human-level performance.
And the weak reasoning ability of the models is one of the important factors for performance failure.

\section{Conclusion}
\label{Conclusion}
In this paper, we propose a novel multi-modal reasoning model named GoT-CQA for challenging CQA task.
The core of GoT-CQA is the compositional reasoning module guided by the interesting graph-of thought (GoT). Note that the GoT transforms the complex question into a operation sequence with several simple operators, and the auto-compositional reasoning pattern is suitable to various type of chart-oriented questions.
Extensive experiments on ChartQA and PlotQA-D show the effectiveness and superiority of GoT-CQA. 
In the future, we intend to improve the GoT-CQA’s ability by exploring more reasonable operators.

{\small
\bibliographystyle{ieee_fullname}
\bibliography{egbib}

\begin{thebibliography}{10}\itemsep=-1pt

\bibitem{brown2020language}
Tom Brown, Benjamin Mann, Nick Ryder, Melanie Subbiah, Jared~D Kaplan, Prafulla Dhariwal, Arvind Neelakantan, Pranav Shyam, Girish Sastry, Amanda Askell, et~al.
\newblock Language models are few-shot learners.
\newblock {\em NeurIPS}, 33:1877--1901, 2020.

\bibitem{chaudhry2020leaf}
Ritwick Chaudhry, Sumit Shekhar, Utkarsh Gupta, Pranav Maneriker, Prann Bansal, and Ajay Joshi.
\newblock Leaf-qa: Locate, encode \& attend for figure question answering.
\newblock In {\em WACV}, pages 3512--3521, 2020.

\bibitem{chen2019figure}
Charles Chen, Ruiyi Zhang, Eunyee Koh, Sungchul Kim, Scott Cohen, Tong Yu, Ryan Rossi, and Razvan Bunescu.
\newblock Figure captioning with reasoning and sequence-level training.
\newblock {\em arXiv preprint arXiv:1906.02850}, 2019.

\bibitem{cheng2023chartreader}
Zhi-Qi Cheng, Qi Dai, and Alexander~G Hauptmann.
\newblock Chartreader: A unified framework for chart derendering and comprehension without heuristic rules.
\newblock In {\em ICCV}, pages 22202--22213, 2023.

\bibitem{devlin2018bert}
Jacob Devlin, Ming-Wei Chang, Kenton Lee, and Kristina Toutanova.
\newblock Bert: Pre-training of deep bidirectional transformers for language understanding.
\newblock {\em arXiv preprint arXiv:1810.04805}, 2018.

\bibitem{huang2023summaries}
Chieh-Yang Huang, Ting-Yao Hsu, Ryan Rossi, Ani Nenkova, Sungchul Kim, Gromit Yeuk-Yin Chan, Eunyee Koh, Clyde~Lee Giles, and Ting-Hao'Kenneth' Huang.
\newblock Summaries as captions: Generating figure captions for scientific documents with automated text summarization.
\newblock In {\em INLG}, pages 80--92, 2023.

\bibitem{jung2017chartsense}
Daekyoung Jung, Wonjae Kim, Hyunjoo Song, Jeong-in Hwang, Bongshin Lee, Bohyoung Kim, and Jinwook Seo.
\newblock Chartsense: Interactive data extraction from chart images.
\newblock In {\em CHI}, pages 6706--6717, 2017.

\bibitem{kafle2018dvqa}
Kushal Kafle, Brian Price, Scott Cohen, and Christopher Kanan.
\newblock Dvqa: Understanding data visualizations via question answering.
\newblock In {\em CVPR}, pages 5648--5656, 2018.

\bibitem{kafle2020answering}
Kushal Kafle, Robik Shrestha, Scott Cohen, Brian Price, and Christopher Kanan.
\newblock Answering questions about data visualizations using efficient bimodal fusion.
\newblock In {\em WACV}, pages 1498--1507, 2020.

\bibitem{kahou2017figureqa}
Samira~Ebrahimi Kahou, Vincent Michalski, Adam Atkinson, {\'A}kos K{\'a}d{\'a}r, Adam Trischler, and Yoshua Bengio.
\newblock Figureqa: An annotated figure dataset for visual reasoning.
\newblock In {\em ICLR}, 2017.

\bibitem{kanthara2022chart}
Shankar Kanthara, Rixie Tiffany~Ko Leong, Xiang Lin, Ahmed Masry, Megh Thakkar, Enamul Hoque, and Shafiq Joty.
\newblock Chart-to-text: A large-scale benchmark for chart summarization.
\newblock {\em arXiv preprint arXiv:2203.06486}, 2022.

\bibitem{kato2022parsing}
Hajime Kato, Mitsuru Nakazawa, Hsuan-Kung Yang, Mark Chen, and Bj{\"o}rn Stenger.
\newblock Parsing line chart images using linear programming.
\newblock In {\em WACV}, pages 2109--2118, 2022.

\bibitem{kim2022ocr}
Geewook Kim, Teakgyu Hong, Moonbin Yim, JeongYeon Nam, Jinyoung Park, Jinyeong Yim, Wonseok Hwang, Sangdoo Yun, Dongyoon Han, and Seunghyun Park.
\newblock Ocr-free document understanding transformer.
\newblock In {\em ECCV}, pages 498--517, 2022.

\bibitem{lee2023pix2struct}
Kenton Lee, Mandar Joshi, Iulia~Raluca Turc, Hexiang Hu, Fangyu Liu, Julian~Martin Eisenschlos, Urvashi Khandelwal, Peter Shaw, Ming-Wei Chang, and Kristina Toutanova.
\newblock Pix2struct: Screenshot parsing as pretraining for visual language understanding.
\newblock In {\em ICML}, pages 18893--18912, 2023.

\bibitem{levy2022classification}
Matan Levy, Rami Ben-Ari, and Dani Lischinski.
\newblock Classification-regression for chart comprehension.
\newblock In {\em ECCV}, pages 469--484, 2022.

\bibitem{li2023weakly}
Hao Li, Jinfa Huang, Peng Jin, Guoli Song, Qi Wu, and Jie Chen.
\newblock Weakly-supervised 3d spatial reasoning for text-based visual question answering.
\newblock {\em TIP}, 32:3367--3382, 2023.

\bibitem{matcha}
Fangyu Liu, Francesco Piccinno, Syrine Krichene, Chenxi Pang, Kenton Lee, Mandar Joshi, Yasemin Altun, Nigel Collier, and Julian Eisenschlos.
\newblock {M}at{C}ha: Enhancing visual language pretraining with math reasoning and chart derendering.
\newblock In {\em ACL}, pages 12756--12770, 2023.

\bibitem{liu2019data}
Xiaoyi Liu, Diego Klabjan, and Patrick NBless.
\newblock Data extraction from charts via single deep neural network.
\newblock {\em arXiv preprint arXiv:1906.11906}, 2019.

\bibitem{2020Multilingual}
Yinhan Liu, Jiatao Gu, Naman Goyal, Xian Li, Sergey Edunov, Marjan Ghazvininejad, Mike Lewis, and Luke Zettlemoyer.
\newblock Multilingual denoising pre-training for neural machine translation.
\newblock In {\em ACL}, pages 726--742, 2020.

\bibitem{luo2021chartocr}
Junyu Luo, Zekun Li, Jinpeng Wang, and Chin-Yew Lin.
\newblock Chartocr: Data extraction from charts images via a deep hybrid framework.
\newblock In {\em WACV}, pages 1917--1925, 2021.

\bibitem{masry2021integrating}
Ahmed Masry and Enamul Hoque.
\newblock Integrating image data extraction and table parsing methods for chart question answering.
\newblock In {\em CVPR}, pages 1--5, 2021.

\bibitem{unichart}
Ahmed Masry, Parsa Kavehzadeh, Xuan~Long Do, Enamul Hoque, and Shafiq Joty.
\newblock Unichart: A universal vision-language pretrained model for chart comprehension and reasoning.
\newblock In {\em EMNLP}, pages 14662--14684, 2023.

\bibitem{masry2022chartqa}
Ahmed Masry, Do~Xuan Long, Jia~Qing Tan, Shafiq Joty, and Enamul Hoque.
\newblock Chartqa: A benchmark for question answering about charts with visual and logical reasoning.
\newblock In {\em Findings of ACL}, pages 2263--2279, 2022.

\bibitem{methani2020plotqa}
Nitesh Methani, Pritha Ganguly, Mitesh~M Khapra, and Pratyush Kumar.
\newblock Plotqa: Reasoning over scientific plots.
\newblock In {\em WACV}, pages 1527--1536, 2020.

\bibitem{qian2021generating}
Xin Qian, Eunyee Koh, Fan Du, Sungchul Kim, Joel Chan, Ryan~A Rossi, Sana Malik, and Tak~Yeon Lee.
\newblock Generating accurate caption units for figure captioning.
\newblock In {\em WWW}, pages 2792--2804, 2021.

\bibitem{rane2021chartreader}
Chinmayee Rane, Seshasayee~Mahadevan Subramanya, Devi~Sandeep Endluri, Jian Wu, and C~Lee Giles.
\newblock Chartreader: Automatic parsing of bar-plots.
\newblock In {\em IRI}, pages 318--325, 2021.

\bibitem{wang2023disavr}
Yaxian Wang, Bifan Wei, Jun Liu, Lingling Zhang, Jiaxin Wang, and Qianying Wang.
\newblock Disavr: Disentangled adaptive visual reasoning network for diagram question answering.
\newblock {\em TIP}, 32:4812--4827, 2023.

\bibitem{2023enhanced}
Mingyang Zhou, Yi Fung, Long Chen, Christopher Thomas, Heng Ji, and Shih-Fu Chang.
\newblock Enhanced chart understanding via visual language pre-training on plot table pairs.
\newblock In {\em Findings of ACL}, pages 1314--1326, 2023.

\end{thebibliography}
}

\end{document}